\documentclass{article}
\usepackage[T1]{fontenc}     
\usepackage{ragged2e}        
\usepackage[hang,flushmargin]{footmisc} 

\usepackage{array}
\newcolumntype{P}[1]{>{\raggedright\arraybackslash}p{#1}}
\usepackage[T1]{fontenc}     
\usepackage{ragged2e}        
\usepackage[protrusion=true,expansion=false]{microtype}  
\usepackage{graphicx}
\usepackage{multirow}
\usepackage{caption}
\usepackage{float}
\usepackage{booktabs}
\usepackage{multicol}
\usepackage{tablefootnote}
\usepackage{textgreek} 
\usepackage{tabularx}        

\usepackage[table,xcdraw]{xcolor}
\usepackage{amsmath,amssymb,amsfonts}
\usepackage{mathtools} 
\usepackage{amsthm}
\usepackage{mathrsfs}
\usepackage{algorithm}
\usepackage{algpseudocode}  
\usepackage{comment}

\usepackage{textcomp}
\usepackage{pifont}
\usepackage{setspace,lineno}
\usepackage{listings}
\usepackage{soul}            
\sethlcolor{yellow}          
\usepackage[title]{appendix}
\usepackage{cite}
\usepackage{manyfoot}
\usepackage{xcolor}

\definecolor{customgreen}{HTML}{00B050}
\definecolor{captionblue}{HTML}{0070C0}  
\newcommand{\cmark}{\textcolor{green}{\ding{51}}} 
\newcommand{\xmark}{\textcolor{red}{\ding{55}}} 
\DeclareCaptionLabelFormat{boldblue}{\textbf{\textcolor{captionblue}{#1 #2}}}
\usepackage{titlesec}
\titleformat{\section}
  {\normalfont\Large\bfseries\color{captionblue}}{\thesection}{1em}{}
\titleformat{\subsection}
  {\normalfont\large\bfseries\color{captionblue}}{\thesubsection}{1em}{}
\titleformat{\subsubsection}
  {\normalfont\normalsize\bfseries\color{captionblue}}{\thesubsubsection}{1em}{}
\usepackage{geometry}
\usepackage{amsmath} 
\allowdisplaybreaks[4]
\usepackage[protrusion=true,expansion=false]{microtype}
\usepackage{hyperref}
\hypersetup{
    colorlinks=true,
    linkcolor=blue,
    citecolor=blue,
    urlcolor=blue
}

\title{DeepAgent: A Dual Stream Multi Agent Fusion for Robust Multimodal Deepfake Detection}

\author{
Sayeem Been Zaman\textsuperscript{1,2,a},
Wasimul Karim\textsuperscript{1,2,3,a}, 
Arefin Ittesafun Abian\textsuperscript{1,2,b,*}, \\
Reem E. Mohamed\textsuperscript{4}, 
Md Rafiqul Islam\textsuperscript{5},
Asif Karim\textsuperscript{5}
Sami Azam\textsuperscript{5,b,*} \\
\small
\textsuperscript{1}Applied Artificial INtelligence and Intelligent Systems (AAIINS) Laboratory, Dhaka 1217, Bangladesh \\
\small
\textsuperscript{2}Department of Computer Science and Engineering, United International University, Dhaka 1212, Bangladesh \\
\small
\textsuperscript{3}Department of Computer Science and Engineering, University of Scholars, Dhaka 1213, Bangladesh\\
\small
\textsuperscript{4}Faculty of Science and Information Technology, Charles Darwin University, Sydney, NSW, Australia \\
\small
\textsuperscript{5}Faculty of Science and Technology, Charles Darwin University, Casuarina, NT 0909, Australia
}

\date{} 
\begin{document}


\justifying
\twocolumn[
\maketitle
\begin{abstract}  
\noindent  The increasing use of synthetic media, particularly deepfakes, is an emerging challenge for digital content verification. Although recent studies use both audio and visual information, most integrate these cues within a single model, which remains vulnerable to modality mismatches, noise, and manipulation. To address this gap, we propose DeepAgent, an advanced multi-agent collaboration framework that simultaneously incorporates both visual and audio modalities for the effective detection of deepfakes. DeepAgent consists of two complementary agents. Agent-1 examines each video with a streamlined AlexNet-based CNN to identify the symbols of deepfake manipulation, while Agent-2 detects audio-visual inconsistencies by combining acoustic features, audio transcriptions from Whisper, and frame-reading sequences of images through EasyOCR. Their decisions are fused through a Random Forest meta-classifier that improves final performance by taking advantage of the different decision boundaries learned by each agent. This study evaluates the proposed framework using three benchmark datasets to demonstrate both component-level and fused performance. Agent-1 achieves a test accuracy of 94.35\% on the combined Celeb-DF and FakeAVCeleb datasets. On the FakeAVCeleb dataset, Agent-2 and the final meta-classifier attain accuracies of 93.69\% and 81.56\%, respectively. In addition, cross-dataset validation on DeepFakeTIMIT confirms the robustness of the meta-classifier, which achieves a final accuracy of 97.49\%, and indicates a strong capability across diverse datasets. These findings confirm that hierarchy-based fusion enhances robustness by mitigating the weaknesses of individual modalities and demonstrate the effectiveness of a multi-agent approach in addressing diverse types of manipulations in deepfakes.
\end{abstract}

\vspace{0.5em}
\noindent \textbf{Keywords:} Deepfake, multi-agent, random forest, meta-classifier
\vspace{1em}
]
{
\renewcommand{\thefootnote}{}
\footnotetext{
\noindent
\textsuperscript{a} Equal Contributions\\
\textsuperscript{b} Equal Supervision\\
\textsuperscript{*} Correspondence:
\begin{tabular}[t]{@{}l@{}}
\href{mailto:aabian191042@bscse.uiu.ac.bd}{aabian191042@bscse.uiu.ac.bd}\\
\href{mailto:sami.azam@cdu.edu.au}{sami.azam@cdu.edu.au}
\end{tabular}%

}

\section{Introduction}
\label{sec1}
The rapid growth of deepfakes has created a critical challenge in ensuring the authenticity and safety of online audio-visual content \cite{hashmi2024understanding, peng2025fairforensics, alsaeedi2025audio}. Deepfakes and related techniques, such as feature editing and lip-syncing, can convincingly mix fake and real footage \cite{peng2025revisiting, han2025hsff}, making them powerful tools for manipulation that endanger individuals and society and cause serious privacy, security and ethical concerns \cite{he2024gazeforensics, liang2023depth, ke2023df}. As these manipulations become more natural and consistent, effective detection requires the analysis of multiple layers of evidence that extend beyond low-level visual cues or basic audio patterns \cite{du2025cad, abbasi2025comprehensive}. Deepfakes can contain fine pixel irregularities, temporal distortions, or inconsistencies between spoken and visible attributes \cite{mao2025leveraging, gao2024temporal, liu2024lips}. These clues often appear unevenly between modalities; some are purely visual, others are embedded in acoustic patterns, and many only emerge when audio, visual, and semantic streams are compared together \cite{du2025cad, zhou2021joint}. Such diverse and cross-modal evidence is difficult to capture reliably using any type of analysis or model alone \cite{hashmi2024understanding, du2025cad}. This highlights the critical need for an automated system that can separately analyze visual and audio artifacts, audio-visual consistency, and then systematically combine these different perspectives. A system designed with specialized agents and a strong decision-level fusion mechanism is essential to handle signals, support interpretability, and improve reliability in real-world and manipulative environments \cite{subudhi2024adaptive}.

Various approaches to deepfake detection have evolved from convolutional neural network (CNN) based methods to transformer-based and multimodal frameworks. Initially, early approaches relied primarily on convolutional and hybrid networks to capture spatial and temporal inconsistencies in facial regions. For example, models such as optical flow (OP) based CNN-LSTM \cite{saikia2022hybrid} and a combination of YOLO-CNN-XGBoost \cite{ismail2021new} have been used to capture spatial and temporal inconsistencies across video frames. Subsequently, more recent developments have shifted toward transformer-based and multimodal architectures that align cross-domain features more effectively. Methods such as CViT \cite{soudy2024deepfake} and ViT-based DeepFake-Adapter \cite{shao2025deepfake}, demonstrate that combining local convolutional extraction with global attention mechanisms can improve deepfake detection accuracy. Furthermore, Multimodal systems such as Spatio-temporal Knowledge Distilled Video vision transformer (STKD-VViT) \cite{usmani2025spatio} utilize both auditory and visual modalities to detect audio-visual mismatches characteristic of deepfakes. Additionally, pretrained models at scale, for instance, Contrastive Language Image Pretraining (CLIP), have been further used and adapted toward cross-domain adaptability using methods such as Category Common Prompt CLIP (C2P-CLIP) \cite{tan2025c2p} focus on cues at the artifact level.

Recent advances \cite{saikia2022hybrid, ismail2021new, soudy2024deepfake, shao2025deepfake, usmani2025spatio, tan2025c2p, sadiq2023deepfake, asuai2025hybrid, raza2022novel, shi2025real, shao2025robust} in deepfake detection have demonstrated improvements in accuracy on benchmark datasets; however, several critical gaps remain unaddressed in current state-of-the-art (SOTA) research. Most existing models rely mainly on unimodal inputs, either visual \cite{saikia2022hybrid, ismail2021new, soudy2024deepfake, shao2025deepfake, tan2025c2p, raza2022novel, shi2025real, shao2025robust} or audio \cite{asuai2025hybrid}, without effectively integrating multimodal cues across vision, audio, and textual semantics. While a few studies \cite{usmani2025spatio} have explored audio-visual fusion, they rely on implicit feature-level fusion that does not explicitly model semantic and lexical consistency between audio and visual contents. Furthermore, existing fusion strategies are typically limited to single stage feature-level or temporal fusion \cite{soudy2024deepfake, shao2025deepfake, usmani2025spatio, sadiq2023deepfake, asuai2025hybrid, shi2025real, shao2025robust}, lacking robust decision-level aggregation mechanisms that can handle conflicting evidence across modalities. Critically, all these approaches do not incorporate collaborative multi-agent architectures capable of adaptive decision-making across different modalities. As a result, they may struggle to maintain performance consistency when applied to diverse datasets.

To address these limitations, we introduce DeepAgent, a novel multi-agent collaborative framework for deepfake detection. In contrast to conventional single-agent detectors, DeepAgent comprises two specialised agents that are trained to focus on different aspects of the manipulation. Agent-1 is a lightweight, memory-efficient CNN derived from AlexNet, designed to capture pixel and frame level visual artifacts in videos. Agent-2 is an audio-visual semantic consistency detector that combines Mel-Frequency Cepstral Coefficient (MFCC)-based acoustic embeddings, Whisper generated speech transcripts, and OCR extracted frame text to quantify lexical and semantic alignment between what is seen and what is heard. Their outputs are fused through a Random Forest meta-classifier, which operates as a decision-level aggregation layer. This collaborative design enables each agent to compensate for the other’s weaknesses: Agent-1 contributes strong visual discrimination, even when audio is noisy or unavailable, while Agent-2 captures cross-modal contradictions in cases where subtle visual artifacts coexist with semantic or temporal misalignments.

The proposed framework is evaluated on three benchmark datasets: Celeb-DF, FakeAVCeleb, and DeepFakeTIMIT. Agent-1 achieves a test accuracy of 94.35\% on the combined Celeb-DF and FakeAVCeleb datasets, demonstrating that a streamlined CNN can still provide competitive visual detection performance. We have included the Celeb-DF dataset alongside the FakeAVCeleb dataset for Agent-1 to provide broader variation in visual manipulations, thereby improving the model’s robustness. Agent-2 attains 93.69\% accuracy on FakeAVCeleb by leveraging multimodal semantic consistency cues. When their predictions are fused, the Random Forest meta-classifier reaches 81.56\% accuracy on FakeAVCeleb and 97.49\% accuracy with an F1-score of 97.52\% on DeepFakeTIMIT, indicating strong cross-dataset generalization. While some transformer-based multimodal baselines report higher performance on individual datasets, DeepAgent offers a balance between accuracy, model complexity, and robustness, particularly in cross-dataset settings where collaborative multi-agent reasoning and decision-level fusion help maintain stable performance across diverse manipulation types and recording conditions.

The contributions of this work can be summarized as follows:
\begin{itemize}
    \item We propose DeepAgent, a dual agent deepfake detection framework that integrates Agent-1, a CNN-based visual module for artifact analysis, with Agent-2, a multimodal module that captures audio-visual inconsistencies using acoustic features, Whisper transcriptions, and OCR-based frame reading.  
    \item We design Agent-1, a lightweight and memory-efficient CNN architecture that achieves high accuracy in image-based deepfake detection while remaining suitable for deployment in constrained environments. 
    \item We develop Agent-2, a multimodal feature extraction and reasoning framework that combines MFCC embeddings, speech transcripts, and visual text to identify semantic and temporal discrepancies characteristic of audio-visual deepfakes.
    \item We implement a Random Forest meta-classifier that fuses the probabilistic outputs of both agents, providing feature and decision level fusion and achieving robust performance across multiple benchmark datasets, especially in cross-dataset validation scenarios.
\end{itemize}

The remainder of the paper is organised as follows. Section \ref{sec2} reviews prior work in deepfake detection. Section \ref{sec3} details the architectures and processing pipelines of Agent-1 and Agent-2, as well as the meta-classification strategy. Section \ref{sec4} describes the datasets, preprocessing, and implementation setup. Section \ref{sec5} presents experimental results, and Section \ref{sec6} provides an in-depth discussion and ablation analysis. Finally, Section \ref{sec7} concludes the paper and outlines future research directions.

\section{Related Works}\label{sec2}
In this section, we review recent advances in deepfake detection and highlight the use of transformer enabled and multimodal frameworks, multi-domain cues, and multi-attention mechanisms, as these have been the recent focus for handling manipulation diversity and generalization challenges.

\subsection{Vision-Based Deepfake Detection Models}
The early research \cite{saikia2022hybrid,ismail2021new,soudy2024deepfake,raza2022novel} in deepfake detection has focused on the visual modalities where convolutional and hybrid deep networks are used to identify traces of manipulation and spatiotemporal inconsistencies between frames.
Saikia et al. \cite{saikia2022hybrid} proposed a spatiotemporal deepfake detection model that integrated CNNs with Long Short Term Memory (LSTM) networks. The model captured inter-frame motion inconsistencies by combining spatial patterns and OP-based temporal features. It achieved an accuracy of 66.26\%, 91.21\%, and 79.49\% on the DFDC, FaceForensics++ (FF++), and Celeb-DF datasets, respectively.  Similarly, Ismail et al. \cite{ismail2021new} proposed YOLO-CNN-XGBoost, a hybrid deep learning system that incorporated YOLO-based face detection, offered deep spatial feature extraction, and XGBoost to provide a final classification. It was tested using the Celeb-DF \& FF++ combined dataset and achieved 90.73\% accuracy, 93.53\% specificity, and 86.36\% F1-score with the benefit of combining deep spatial representations and using an ensemble of decision boundaries to produce strong detection and lower false positives.

Shi et al. \cite{shi2025real} introduced the Real Face Foundation Representation Learning (RFFR) approach, which aimed to learn a generalized representation from large collections of real face images. The method identified abnormal artifacts that fall outside this learned distribution and achieved an accuracy of 88.98\% on the Celeb-DF dataset. Further improving this, Soudy et al. \cite{soudy2024deepfake} proposed a visual deepfake detection framework with CNNs using the Convolutional Vision Transformers (CViTs) to maximize the representation of multi-region features. Their model used parallel CNN and CViT modules to analyze the face, eyes, and nose components with an accuracy of 97\% and 85\%, respectively, in CNN-based and CViT-based detections on the FF++ and DFDC datasets. In another study, Raza et al. \cite{raza2022novel} proposed Deepfake Predictor (DFP), a deep learning architecture that used VGG16 with a custom CNN architecture, which reached 94\% accuracy and 95\% precision on a dataset of real and fake human faces. Collectively, these vision-based approaches offer useful baselines for image and video forgery detection


\subsection{Audio and Multi-Modal Deepfake Detection Models}
Recent research has gone beyond the vision-based deepfake detection to include audio and multimodal fusion methods to detect cross-domain inconsistencies between visual and auditory information \cite{asuai2025hybrid,usmani2025spatio}.
Asuai et al. \cite{asuai2025hybrid} proposed a hybrid CNN-LSTM network that can be used to detect synthetic speech through a mixture of Mel Frequency Cepstral Coefficients (MFCCs) and spectrogram. The framework also achieved an accuracy of 94.7\% and an AUC of 97.3\% on the Fake-or-Real (FoR) dataset, showing that it is robust and can be generalized across different benchmarks. Similarly, Usmani et al. \cite{usmani2025spatio} proposed STKD-VViT, a transformer-based multimodal framework that consistently analyzed audio and video streams using spatio-temporal patterns. They applied the knowledge distillation in the model to reduce the model's parameters, transferring the knowledge from a large complex model to a lightweight small model, without compromising on accuracy. It was tested on the FakeAVCeleb dataset and got a fusion-based accuracy of 96\%.

\begin{table*}[ht!]
\centering
\scriptsize
\caption{Comparative analysis of existing deepfake detection methods and DeepAgent, highlighting modality, architecture, fusion strategy, semantic consistency reasoning, and cross-dataset evaluation.}
\begin{tabular}{
>{\centering\arraybackslash}p{2.3cm}
>{\centering\arraybackslash}p{1.3cm}
>{\centering\arraybackslash}p{2.3cm}
>{\centering\arraybackslash}p{2.3cm}
>{\centering\arraybackslash}p{1.7cm}
>{\centering\arraybackslash}p{1.7cm}
>{\centering\arraybackslash}p{2.2cm}
>{\centering\arraybackslash}p{1.6cm}
}
\toprule
\textbf{Study} & \textbf{Modality} & \textbf{Architecture} & \textbf{Dataset(s)} & \textbf{Acc. (\%)} & \textbf{Fusion} & \textbf{AV Consistency} & \textbf{Cross-dataset} \\
\midrule
Saikia et al. \cite{saikia2022hybrid} & Video & CNN + LSTM & DFDC, FF++, Celeb-DF & 66.26 / 91.21 / 79.49 & None & None & No \\
Ismail et al. \cite{ismail2021new} & Video & YOLO + CNN + XGB & Celeb-DF, FF++ & 90.73 & None & None & No \\
Soudy et al. \cite{soudy2024deepfake} & Video & CNN + CViT & FF++, DFDC & 97.0 / 85.0 & Feature-level & None & No \\
Shao et al. \cite{shao2025deepfake} & Video & ViT + Adapters & FF++ & 96.52 & Feature-level & None & No \\
Usmani et al. \cite{usmani2025spatio} & Audio + Video & STKD-VViT & FakeAVCeleb & 96.0 & Feature-level & Implicit & No \\
Tan et al. \cite{tan2025c2p} & Video & CLIP-based & GenImage & 95.80 & None & None & Not reported \\
Sadiq et al. \cite{sadiq2023deepfake} & Text & CNN + FastText & TweepFake & 93.0 & Feature-level & None & No \\
Asuai et al. \cite{asuai2025hybrid} & Audio & CNN + LSTM & FoR & 94.70 & Feature-level & Implicit (Audio only) & No \\
Raza et al. \cite{raza2022novel} & Image & VGG16 + CNN & Real/Fake Face & 94.0 & None & None & No \\
Shi et al. \cite{shi2025real} & Video & ViT (RFFR) & Celeb-DF, DFDC & 88.98 / 67.84 & Feature-level & None & No \\
Shao et al. \cite{shao2025robust} & Video & SeqFakeFormer++ & Seq-DeepFake & 72.81 & Temporal & None & No \\
Ours (DeepAgent) & Audio + Video + Text & CNN + DNN + RF & Celeb-DF, FakeAVCeleb, DFTIMIT & 81.56 / 97.49 & Feature + Decision & Explicit AV lexical \& semantic & Yes \\
\bottomrule
\end{tabular}
\label{tab:gap_analysis}
\end{table*}

\subsection{Generalization and Sequential Manipulation Understanding}
In recent studies \cite{tan2025c2p, shao2025deepfake, shao2025robust, sadiq2023deepfake} about deepfake detection, there has been a focus on model generalization, interpretability, and sequential reasoning based on hybrid architectures and transformer-based aligning feature strategy. To begin with, Tan et al. \cite{tan2025c2p} proposed C2P-CLIP, a category prompted CLIP model designed to address domain generalization. It inserted conceptual prompts into the text encoder and achieved an accuracy of 95.8\% on the Genimage dataset. This approach demonstrated that cross-domain performance can be improved using semantic prompt adaptation. However, C2P-CLIP focuses primarily on visual domain generalization and does not explicitly incorporate audio features or cross-modal semantic reasoning. 

Similarly, Shao et al. \cite{shao2025deepfake} presented DeepFake-Adapter, a parameter-efficient framework for Vision Transformer (ViT) adaptation that used global and local adapter modules to extract fine grained and semantic patterns of forgery. It achieved an accuracy of 96.52\% on the FF++ dataset with ViT using the neural texture technique. Despite its parameter-efficient design, DeepFake-Adapter remains a single-modal visual detector and does not explicitly address misalignment between what is seen and what is heard. In another study, Shao et al. \cite{shao2025robust} introduced Seq-DeepFake and its enhanced version, SeqFakeFormer++, that reframed deepfake detection as a sequence modeling problem, allowing for detecting the presence and trajectory of manipulations in a video. Their method achieved an accuracy of 72.81\% using a ViT-based model in the Seq-DeepFake dataset. By reframing the task as image to sequence prediction, they achieved temporal transparency, moving beyond simple binary classification outcomes. Moreover, Sadiq et al. \cite{sadiq2023deepfake} proposed a CNN-based deep learning model that combines FastText word embeddings to identify deepfake tweets generated by bots on social media. They constructed the model by preprocessing the TweepFake dataset and converting the textual data into FastText embeddings, achieving an accuracy of 93\%. 

Our proposed approach addresses several critical limitations found in existing deepfake detection studies \cite{saikia2022hybrid, ismail2021new, soudy2024deepfake, shao2025deepfake, usmani2025spatio, tan2025c2p, sadiq2023deepfake, asuai2025hybrid, raza2022novel, shi2025real, shao2025robust} by introducing key insights and solutions as presented in Table \ref{tab:gap_analysis}. Unlike traditional methods that rely primarily on single types of input, either visual \cite{saikia2022hybrid, ismail2021new, soudy2024deepfake, shao2025deepfake, tan2025c2p, raza2022novel, shi2025real, shao2025robust} or audio \cite{asuai2025hybrid} alone, our proposed framework, DeepAgent, effectively integrates multimodal cues across vision, audio, and textual semantics to identify deepfakes. In contrast to earlier approaches that use only early or late feature-level fusion \cite{soudy2024deepfake, shao2025deepfake, usmani2025spatio, sadiq2023deepfake, asuai2025hybrid, shi2025real}, DeepAgent employs a two-stage fusion strategy that combines both feature-level and decision-level reasoning and this allows the system to align conflicting evidence across modalities. Additionally, the framework explicitly models semantic and lexical consistency between audio and visual contents, rather than relying on implicit feature-level fusion used in previous audio-visual approaches \cite{usmani2025spatio, asuai2025hybrid}. Importantly, none of these studies include collaborative multi-agent systems that can adapt their decisions across different modalities. DeepAgent directly addresses these gaps that combines a lightweight visual detector (Agent-1) with an audio–visual semantic consistency agent (Agent-2) and a Random Forest meta-classifier that operates at the decision level, and enables both feature-level and decision-level fusion, supporting robust cross-dataset performance.

\section{Methodology: DeepAgent}\label{sec3}
Our proposed DeepAgent framework is a multi-agent deepfake detection system designed utilizing both visual patterns and audio-visual inconsistencies typically present in manipulated videos. This framework can be applied across several domains, from social media and online communication platforms to digital forensics and security sensitive environments, where reliable deepfake detection is increasingly necessary. The methodology is structured into two primary components: Agent-1, which focuses on the visual features, and Agent-2 applies the multimodal patterns through a combination of frame reading outputs, audio features, and transcriptions. The agents are trained to categorize the real and fake content independently, thus acquiring complementary patterns of decisions. 

In order to further improve the detection performance, we have used a meta-classification approach that involves the combination of the probabilistic outputs of both agents with the help of a Random Forest classifier. This ensemble approach ensures that the final prediction is informed by both spatial and semantic inconsistencies, while increasing the flexibility against deepfakes and modality specific artifacts. This section provides a detailed description of the internal architectures, feature extraction pipelines, and fusion strategy. Figure \ref{fig:deepagent} illustrates the architecture of DeepAgent.

\begin{figure*}[ht!]
\centering
\includegraphics[scale=0.198]{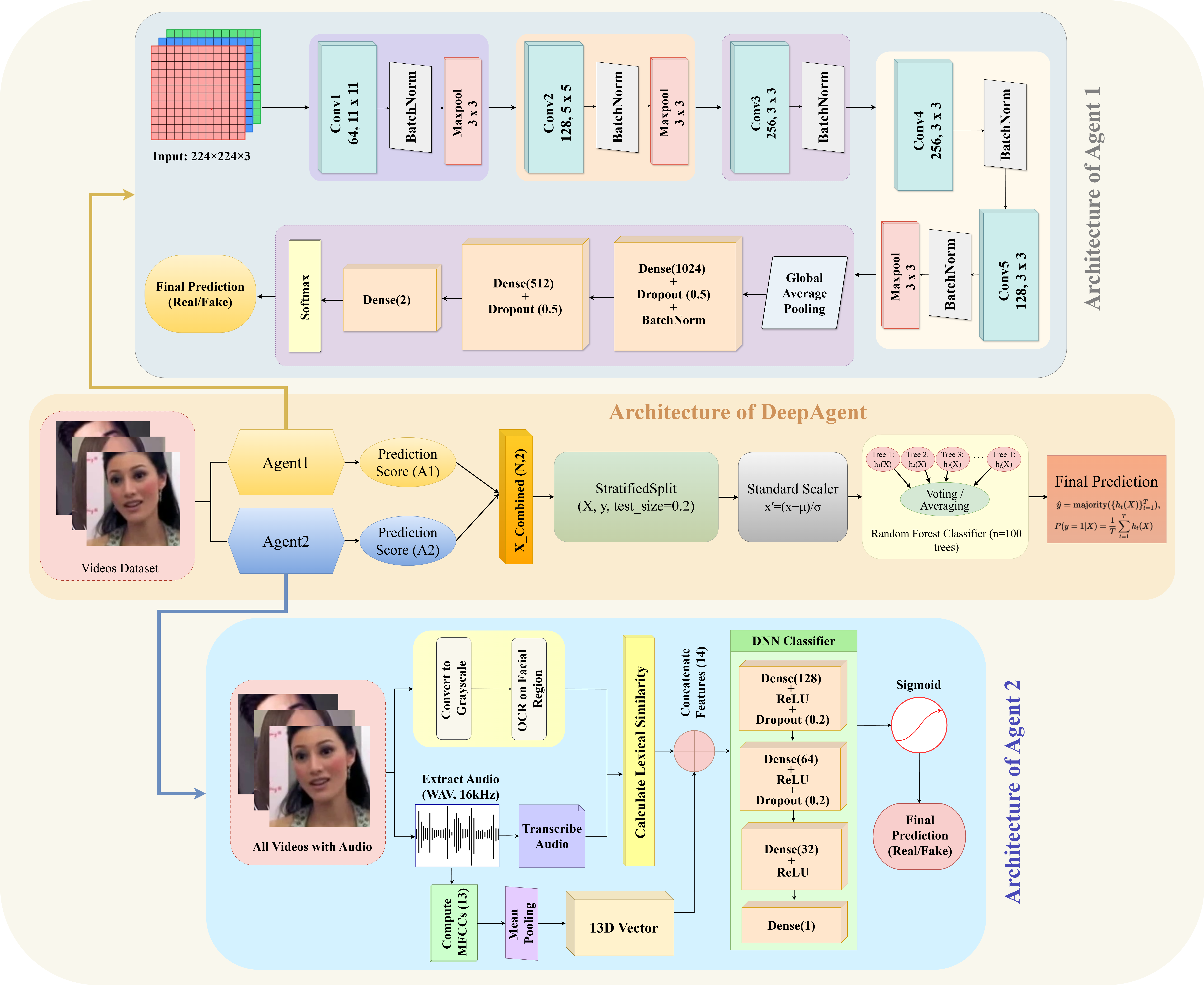}
\caption{Overview of the proposed DeepAgent architecture, which integrates a lightweight CNN, an audio-visual semantic consistency detector using MFCCs, Whisper-based transcription, and OCR-based frame reading, and a Random Forest meta-classifier.}
\label{fig:deepagent}
\end{figure*}

\subsection{Agent-1: Visual-Based Detector}\label{sec32}
We have trained Agent-1 to detect visual patterns that specify a deepfake through a thorough analysis of the details present in each video. This process is supported by a memory-efficient CNN derived from the AlexNet architecture \cite{krizhevsky2012imagenet}. AlexNet's shallow architecture naturally extracts low-level texture and edge features, which are suitable for detecting pixel-level deepfake artifacts while maintaining computational efficiency. The architecture of Agent-1 has been optimized to reduce computational complexity while preserving strong discriminative capability, ensuring accurate and efficient detection performance. Figure \ref{fig:agent1} represents the architecture of Agent-1.

\begin{figure*}[ht!]
\centering
\includegraphics[scale=0.48]{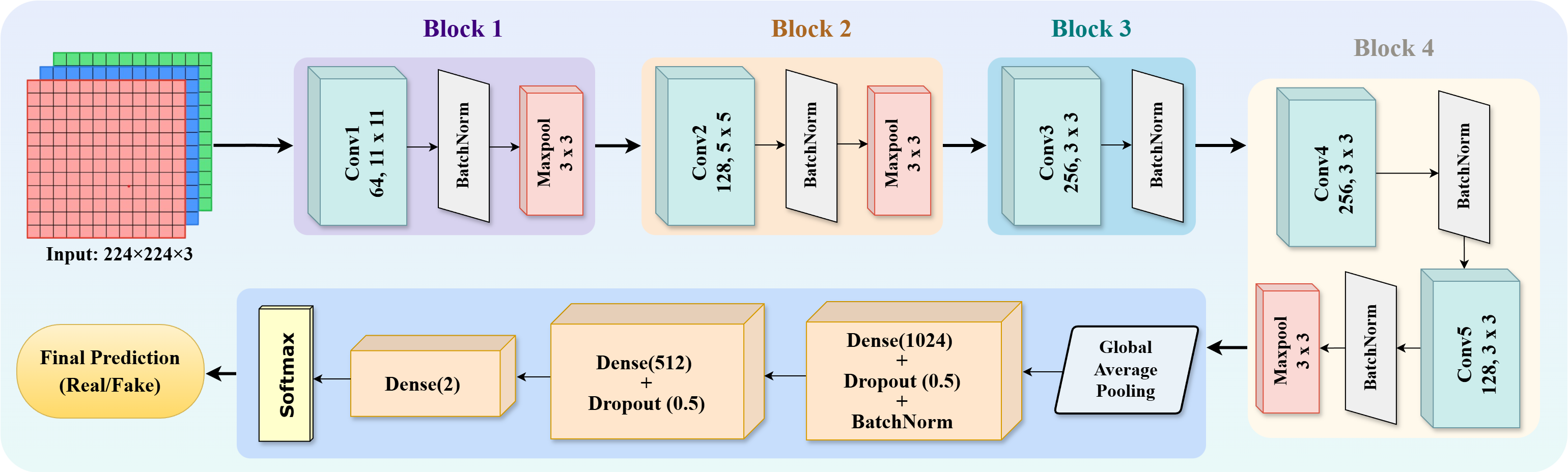}
\caption{The architecture of Agent-1, a streamlined AlexNet-based CNN model with convolutional, batch normalization, pooling, dense, and dropout layers for real/fake prediction.}
\label{fig:agent1}
\end{figure*}
First, the Convolutional Feature Extraction Module converts the input step by step to the higher-level representations by a series of five hierarchical convolutional blocks. Each block captures local spatial information, whereas the ReLU activation function maintains significant non-linear associations in the information. Following this, the mechanism of Normalization and Pooling runs the batch normalization process to stabilize the intermediate level feature distributions and uses max pooling to eliminate unnecessary spatial information to enhance the stability of the model to translation and other visual changes. 

Moreover, the Global Average Pooling (GAP) Representation summarizes the spatial responses into a few compact descriptors, which basically reduces the complexity of the parameters without global semantic loss of information. Furthermore, the Regularized Dense Layers reduce such descriptors into high-level embeddings by using both dropout and batch normalization to prevent overfitting and promote high generalizability. Finally, interpreting such embeddings using the Softmax Decision Module, categorical cross-entropy loss is used, and the Adam optimizer is utilized, both of which enable the optimization mechanism to optimize the network parameters. A combination of these components creates a specialized visual analysis agent that is able to encode spatial features in video efficiently and be robust and compact.

\subsubsection{Convolutional Feature Extraction Module}
We have constructed a Convolutional Feature Extraction Module built on a five block convolutional neural network, which uses sequential convolutional filters, batch normalization, and max pooling to learn spatial representations \cite{krizhevsky2012imagenet}. The backbone of our model is a sequence of five convolutional blocks, each consisting of convolutional filters followed by batch normalization and, with max pooling applied as needed. The first layer applies $64$ filters of size $11 \times 11$ with a stride of $4$, capturing low-level spatial structures. Successive layers increase the depth of representation, progressing through $128$, $256$, and $256$ filters before reducing to $128$ filters in the final block. The network follows an AlexNet-style design with five convolutional blocks followed by three fully connected layers. Each convolution uses a ReLU activation, and batch normalization is applied after every convolutional layer. Max pooling is used after selected blocks to reduce spatial resolution.

The convolution operation for an input tensor $I$ and kernel $K$ is mathematically as defined in Equation \ref{eq:1}:
\begin{equation}
O_{i,j,d} = \sum_{m=0}^{k-1} \sum_{n=0}^{k-1} \sum_{c=0}^{D_{in}-1} I_{i+m, j+n, c} \cdot K_{m,n,c,d} + b_d
\label{eq:1}
\end{equation}
where $O_{i,j,d}$ denotes the output feature map at spatial location $(i, j)$ and depth channel $d$, $m$ and $n$ represent the spatial indices within the convolution kernel window, $I_{i+m, j+n, c}$ represents the input pixel value at position $(i+m, j+n)$ in channel $c$, $K_{m,n,c,d}$ is the kernel weight connecting input channel $c$ to output channel $d$, $b_d$ is the bias term associated with the $d$-th output channel, $D_{in}$ is the total number of input channels, and $k \times k$ defines the spatial size of the convolution kernel.

\subsubsection{Normalization and Pooling Mechanism}
To stabilize the network's training and improve feature robustness, we have employed batch normalization after each convolution and apply max pooling to reduce dimensionality \cite{ioffe2015batch}. This step ensures that inputs to each layer have zero mean and unit variance, thereby accelerating convergence. To achieve dimensionality reduction and enhance translational invariance, max pooling operations are applied after selected convolutional layers. The pooling function is defined in Equation \ref{eq:3}:
\begin{equation}
P_{i,j,d} = \max_{0 \leq m < p,0 \leq n < p} O_{i+m, j+n, d}
\label{eq:3}
\end{equation}
where $P_{i,j,d}$ represents the pooled output value at spatial location $(i, j)$ and depth channel $d$, $O_{i+m, j+n, d}$ denotes the activation value from the convolutional feature map within the pooling window, $p$ indicates the spatial dimension (height and width) of the pooling window, and the indices $m,n$ are integer offsets that run over the $p\times p$. The $\max$ operation selects the highest activation within each $p\times p$ region, effectively preserving the most salient feature responses while reducing spatial resolution and improving translation invariance.

\subsubsection{Global Average Pooling Representation}
Instead of flattening feature maps into high dimensional fully connected layers, the architecture uses GAP to reduce each feature map into a single scalar by averaging spatial dimensions \cite{he2016deep}. This design choice lowers the number of trainable parameters and improves memory efficiency, as described in Equation \ref{eq:4}:
\begin{equation}
    g_d = \frac{1}{H \cdot W} \sum_{i=1}^{H} \sum_{j=1}^{W} O_{i,j,d}
\label{eq:4}
\end{equation}
where $g_d$ denotes the pooled descriptor for the $d$-th channel, $O_{i,j,d}$ represents the activation value at spatial location $(i, j)$ within channel $d$, $H$ is the height of the feature map, and $W$ is its width.

\subsubsection{Regularized Dense Layers}
We have applied a fully connected stage consisting of two regularized dense layers with 1024 and 512 neurons, respectively, where dropout and batch normalization are used to stabilize learning and map the extracted features to the final softmax classification output \cite{taeb2022comparison}. Dropout with a rate of $0.5$ is applied between layers to prevent overfitting by randomly deactivating units during training. Batch normalization is also employed to maintain numerical stability.

The final classification layer employs a softmax activation, which maps the network outputs to class probabilities. For $C$ classes, the probability of class $i$ is given by Equation~\ref{eq:5}:
\begin{equation}
    \hat{y}_i = \frac{e^{z_i}}{\sum_{j=1}^{C} e^{z_j}}
\label{eq:5}
\end{equation}
where $\hat{y}_i$ represents the predicted probability for class $i$, the summation index $j$ ranges over all classes, $z_i$ denotes the preactivation value corresponding to class $i$, and $C$ is the total number of output classes.

\subsubsection{Optimization Mechanism}
We have designed an optimization mechanism that minimizes the categorical cross-entropy loss and updates model parameters using the Adam algorithm, thereby employing adaptive first and second moment gradient estimates \cite{wang2022comprehensive}. The training process optimizes the model parameters by minimizing the categorical cross-entropy loss, which is defined in Equation \ref{eq:6}:
\begin{equation}
    \mathcal{L} = -\sum_{i=1}^{C} y_i \log(\hat{y}_i)
\label{eq:6}
\end{equation}
where $y_i$ represents the ground truth label for class $i$ and $\hat{y}_i$ denotes the predicted probability, as obtained from the softmax function:

Parameter updates are carried out using the Adam optimization algorithm, which adaptively adjusts the learning rate for each parameter based on first and second moment estimates of the gradients. The update rule for Adam is presented in Equation \ref{eq:7}:
\begin{equation}
    \theta_{t+1} = \theta_t - \eta \cdot \frac{\hat{m}_t}{\sqrt{\hat{v}_t} + \epsilon}
\label{eq:7}
\end{equation}
where $\theta_t$ represents the model parameters at iteration $t$, $\theta_{t+1}$ represents the updated parameters after applying the optimization step, $\hat{m}_t$ and $\hat{v}_t$ are the bias corrected first and second moment estimates of the gradient, respectively, and $\eta$ is the learning rate.

\subsection{Agent-2: Audio-Visual Semantic Consistency Detector}\label{sec33}
We have developed Agent-2 to detect possible manipulations in videos by analyzing mismatches between the audio and visual elements \cite{bohacek2024lost}. Unlike systems that rely only on visual data, Agent-2 also uses speech-related meaning and text information from video frames and helps it more easily identify the semantic and sound-based differences often found in deepfakes \cite{li2024zero}. The detection algorithm is structured into the following three phases: (i) audio-visual feature extraction, (ii) constructing multimodal features, and (iii) binary classification through a neural network. Figure \ref{fig:agent2} illustrates the structure of Agent-2 diagrammatically.

\begin{figure*}[ht!]
\centering
\includegraphics[scale=0.5]{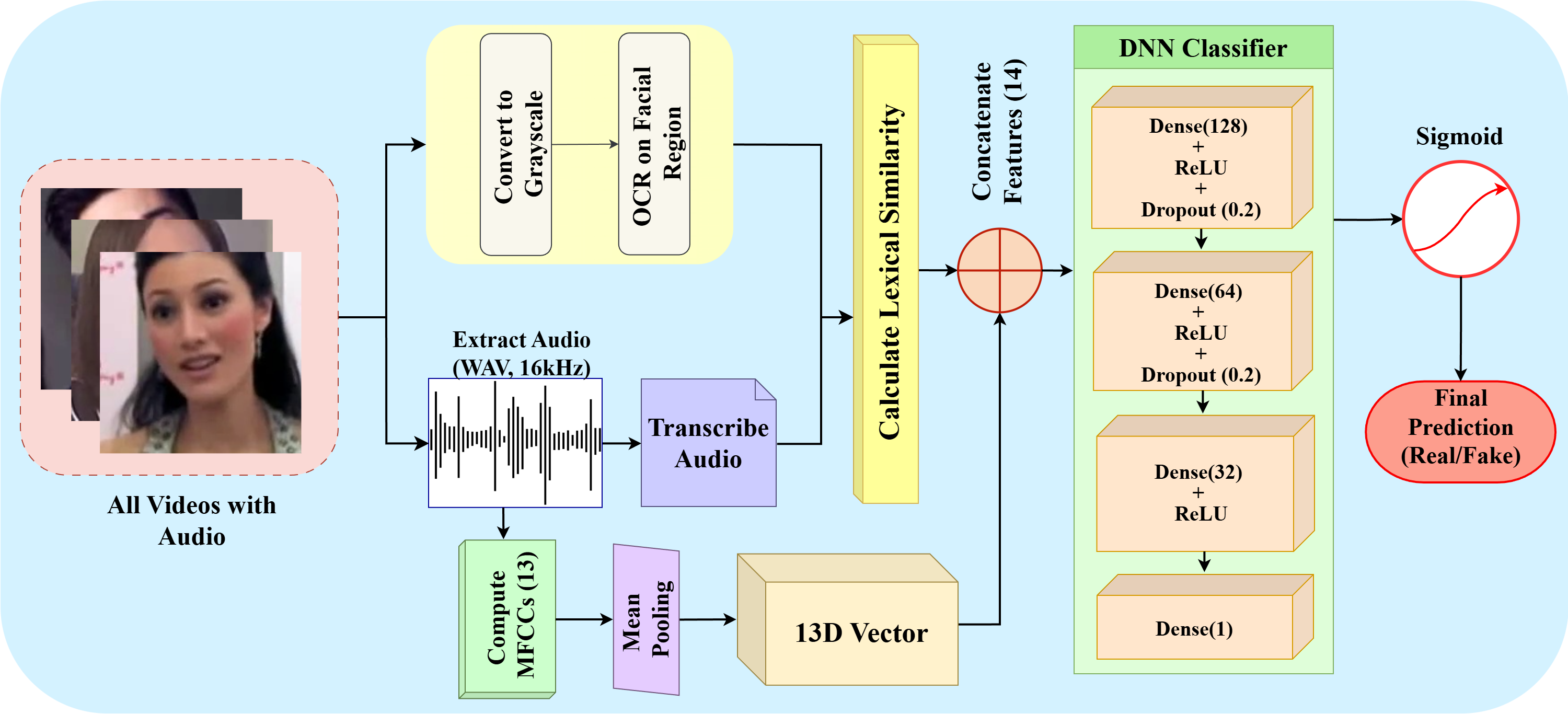}
\caption{The architecture of Agent-2, an audio-visual semantic consistency detector that fuses MFCC-based acoustic features, Whisper generated transcripts, and OCR-based frame text, incorporating lexical similarity and a DNN classifier with dense layers and ReLU activation, followed by a sigmoid output to predict real or fake.}
\label{fig:agent2}
\end{figure*}

\subsubsection{Audio-Visual Feature Extraction Module}
We have developed the Audio-Visual Feature Extraction Module that processes both the audio and visual streams of each video, computes MFCCs from the resampled audio waveform, and extracts textual cues from videos via OCR; the resulting features are then combined through a cross-modal lexical similarity measure \cite{li2024zero}. Intuitively, MFCCs summarize how much energy the signal has in different perceptual frequency bands over time. We first compute a short time Fourier transform (STFT) of the audio, which gives us a spectrum of frequencies for each small time window (e.g., every 20–25 ms). Instead of working with hundreds of narrow frequency bins, we group them into a smaller number of Mel-spaced bands that more closely reflect how humans perceive pitch: more resolution at low frequencies and less at very high frequencies. For each time window and each Mel band, we then sum the spectral energy inside that band. This gives us a compact description of the audio that is robust to small local variations while preserving the global timbre and prosody patterns important for deepfake detection.

Let $V$ denote an input video, which is decomposed into its constituent audio signal $A(t)$ and visual frame sequence $\{F_i\}_{i=1}^N$. The extracted audio signal $A(t)$ is uniformly resampled to a fixed sampling rate $f_s = 16\,\text{kHz}$ to ensure consistency across samples. From the resampled waveform $A(t)$, MFCCs are computed. If $\mathcal{S}(f, \tau)$ denotes the short time magnitude spectrum obtained from windowed Fourier analysis over frame index $\tau$, the Mel-scaled filterbank energies are computed using the following Equation \ref{eq:8}:
\begin{equation}
E_m(\tau) = \sum_{f} H_m(f) \cdot |\mathcal{S}(f, \tau)|^2
\label{eq:8}
\end{equation}
where $E_m(\tau)$ represents the energy of the $m$-th Mel filter at frame $\tau$, $H_m(f)$ is the triangular weighting function corresponding to the $m$-th Mel filter. $f$ denotes the frequency bin index in the short time Fourier transform.

Equation~\ref{eq:9} then applies a discrete cosine transform (DCT) to these log energies. This step decorrelates the bands and compresses most of the useful information into the first 13 coefficients, which we use as our MFCC feature vector. The MFCC for coefficient index $c$ and time frame $\tau$ is computed as shown in Equation~\ref{eq:9}:
\begin{equation}
M_{c,\tau} = \sum_{m=1}^{M} \log(E_m(\tau)) \cdot \cos\left[\frac{\pi c (m - 0.5)}{M}\right]
\label{eq:9}
\end{equation}
where $M_{c,\tau}$ denotes the MFCC value corresponding to coefficient index $c$ and time frame $\tau$; 
$E_m(\tau)$ represents the Mel-filterbank energy for the $m$-th filter at frame $\tau$, $M$ is the total number of Mel filters (set to $M=13$ in this configuration), $c$ is the cepstral coefficient index, $\tau$ denotes the time frame index, and $m$ is the Mel filter index. In the equation, $\pi$ is the mathematical constant pi and $\cos(\cdot)$ denotes the cosine function applied to the argument.

The temporal mean of each coefficient is then computed to obtain the final 13-dimensional audio embedding, as given in Equation~\ref{eq:10}:
\begin{equation}
\mathbf{a} = \frac{1}{T} \sum_{\tau=1}^T \mathbf{M}_{:, \tau} \in \mathbb{R}^{13}
\label{eq:10}
\end{equation}

where $T$ is the total number of time frames, and $\mathbf{M}_{:, \tau}$ denotes the MFCC vector at frame $\tau$. The resulting embedding $\mathbf{a} \in \mathbb{R}^{13}$ represents a compact, fixed length descriptor capturing the average spectral characteristics across all frames.

Each selected frame $F_j$ is then converted to grayscale intensity space $G_j(u,v)$. The formulation is provided in Equation \ref{eq:11}:
\begin{equation}
G_j(u,v) = 0.299 R_j(u,v) + 0.587 G_j(u,v) + 0.114 B_j(u,v)
\label{eq:11}
\end{equation}
where $(u,v)$ represents the spatial pixel coordinates, and $(R_j, G_j, B_j)$ correspond to the red, green, and blue color channel intensities of the $j$-th frame.

The resulting frames are processed using an OCR engine, which extracts textual information. This step produces a set of recognized word tokens $\mathcal{W}_v$, encapsulating both visible text and inferred frame-level content.

The speech signal $A(t)$ is also transcribed into a string sequence $T_a$ from which the set of word tokens detected $\mathcal{W}_a$ is extracted.

A cross-modal lexical similarity score, denoted as $s$, is computed by measuring the normalized size of the intersection between $\mathcal{W}_a$ and $\mathcal{W}_v$, as presented in Equation \ref{eq:12}:
\begin{equation}
s = \frac{|\mathcal{W}_a \cap \mathcal{W}_v|}{\max(|\mathcal{W}_a|, 1)}
\label{eq:12}
\end{equation}
This scalar provides a measure of semantic correspondence between the speech and the facial actions.

\subsubsection{Multi-Modal Feature Construction}
We construct a compact multimodal feature vector by combining the 13-dimensional vector with a cross-modal lexical similarity score, ensuring a fixed 14-dimensional representation for each video \cite{zhou2021joint, schindler2020unsupervised}.
The final multimodal representation $\mathbf{x} \in \mathbb{R}^{14}$ is obtained by concatenating the 13-dimensional MFCC mean vector $\mathbf{a}$ with the lexical similarity score $s$ and the formulation is presented in Equation \ref{eq:13}:
\begin{equation}
\mathbf{x} = \begin{bmatrix} \mathbf{a} \\ s \end{bmatrix} \in \mathbb{R}^{14}
\label{eq:13}
\end{equation}

When speech transcription or audio feature extraction is not working out for some instances, the unavailable pieces of $\mathbf{a}$ are replaced by zero, and a zero centered default for $s$ is used, thereby maintaining a fixed dimensional structure for every sample. This vector contains additional phonetic and semantic content, maintaining spectral features of speech along with cross-modal consistency signals that are taken directly from the input video $V$.

\subsubsection{Binary Decision Network}

The constructed feature vector $\mathbf{x}$ is input to a fully connected deep neural network (DNN) configured for binary classification. Let $\hat{y} = f(\mathbf{x}; \theta)$ denote the output of the network parameterized by weights $\theta$, where $\hat{y} \in [0, 1]$ represents the predicted probability of the video being fake.

The structure of the network is comprised of $L = 4$ layers: one input layer of $128$ ReLU activated neurons, two hidden layers with $64$ and $32$ units respectively, and one output sigmoid activated unit for binary classification. We apply dropout layers with a dropout rate of $p = 0.2$ after the first and second layers to prevent overfitting. The forward propagation process of the network is mathematically formulated and described in Equations~\ref{eq:14}--\ref{eq:19}:

\begin{align}
\mathbf{h}_1 &= \text{ReLU}(\mathbf{W}_1 \mathbf{x} + \mathbf{b}_1), \label{eq:14} \\
\mathbf{h}_1' &= \text{Dropout}(\mathbf{h}_1, p), \label{eq:15} \\
\mathbf{h}_2 &= \text{ReLU}(\mathbf{W}_2 \mathbf{h}_1' + \mathbf{b}_2), \label{eq:16} \\
\mathbf{h}_2' &= \text{Dropout}(\mathbf{h}_2, p), \label{eq:17} \\
\mathbf{h}_3 &= \text{ReLU}(\mathbf{W}_3 \mathbf{h}_2' + \mathbf{b}_3), \label{eq:18} \\
\hat{y} &= \sigma(\mathbf{w}_4^\top \mathbf{h}_3 + b_4). \label{eq:19}
\end{align}

where $\mathbf{x}$ denotes the input vector, $\mathbf{W}_1$, $\mathbf{W}_2$, and $\mathbf{W}_3$ are the weight matrices for each corresponding layer, and $\mathbf{b}_1$, $\mathbf{b}_2$, and $\mathbf{b}_3$ represent the bias vectors. The hidden activations $\mathbf{h}_1$, $\mathbf{h}_2$, and $\mathbf{h}_3$ correspond to the outputs of each ReLU-activated layer, while $\mathbf{h}_1'$ and $\mathbf{h}_2'$ denote the activations after applying dropout with probability $p$. The final prediction $\hat{y}$ is obtained by applying the sigmoid activation function $\sigma$ to the linear combination of the last hidden representation $\mathbf{h}_3$ using output weights $\mathbf{w}_4$ with transpose $\top$ and bias $b_4$.

The network is optimized using the binary cross-entropy loss function, which is presented in Equation \ref{eq:20}:

\begin{equation}
\mathcal{L}(\hat{y}, y) = -y \log(\hat{y}) - (1 - y) \log(1 - \hat{y})
\label{eq:20}
\end{equation}

where $\mathcal{L}(\hat{y}, y)$ represents the total loss, $\hat{y}$ is the predicted probability of the positive class, and $y \in {0,1}$ denotes the ground truth binary label.

Our proposed architecture enables Agent-2 to learn of complicated intermodality relationships and detect forged content from individual stream abnormalities, as well as intermodality alignment. Fusing features from the acoustic, textual, and semantic spaces makes Agent-2 highly resistant to high fidelity deepfakes that have realism in one modality, yet fail for cross-modality coherence.

\subsection{Meta-Classification and Fusion Strategy}
We have developed a final decision making phase that integrates both Agent-1 and Agent-2 into a unified classification framework \cite{rahman2026self}. Through its two-phased structure, this system makes it feasible to utilize complementary discriminative information and offset one another's failures of the detectors. The complete meta-classification and fusion strategy is outlined in Algorithm \ref{alg:meta_fusion}.

\subsubsection{Multi-Agent Fusion Mechanism}
To achieve a more reliable decision, the outputs of both agents are combined through a multi-agent fusion strategy \cite{mungoli2023adaptive}. This approach is designed to integrate different cues extracted from visual and audio modalities.
Let $\mathcal{D} = \{(V_i, y_i)\}_{i=1}^N$ denote the dataset, where $V_i$ is the $i$-th video sample and $y_i \in \{0,1\}$ is its ground truth label ($0$ = real, $1$ = fake).  

For each video $V_i$, Agent-1 processes videos $\{F_{i,j}\}_{j=1}^{n_i}$ at regular intervals of frames (as defined in Section \ref{sec32}), and Agent-1 assigns a prediction score $p^{(1)}_{i,j} \in [0,1]$ to each frame. Here, $p^{(1)}_{i,j}$ indicates the confidence level of Agent-1 for the $j$-th frame of video $V_i$. 

To obtain the overall prediction for the entire video, the frame-level scores are averaged to produce an aggregated score $\hat{y}^{(1)}_i$, as shown in Equation \ref{eq:21}:
\begin{equation}
\hat{y}^{(1)}_i = \frac{1}{n_i} \sum_{j=1}^{n_i} p^{(1)}_{i,j}
\label{eq:21}
\end{equation}
where $\hat{y}^{(1)}_i$ denotes the final prediction score for video $V_i$, $\sum_{j=1}^{n_i}$ represents the summation over all frames, and the factor $\frac{1}{n_i}$ normalizes the sum by the total number of frames.

Agent-2 processes the same video $V_i$ by extracting its 14-dimensional multimodal feature vector $\mathbf{x}_i$ (as defined in Section \ref{sec33}), and outputs a scalar probability as illustrated in Equation \ref{eq:22}:
\begin{equation}
\hat{y}^{(2)}_i = f_{\theta_2}(\mathbf{x}_i) \in [0,1]
\label{eq:22}
\end{equation}
where $f_{\theta_2}$ is the trained binary classifier of Agent-2.  

The outputs from both agents are concatenated to form a two-dimensional meta-feature vector, as expressed in Equation \ref{eq:23}:
\begin{equation}
\mathbf{z}_i = \begin{bmatrix} \hat{y}^{(1)}_i \\ \hat{y}^{(2)}_i \end{bmatrix} \in \mathbb{R}^2
\label{eq:23}
\end{equation}
This vector encodes the independent judgments of the visual and multimodal detectors for each video.

\begin{algorithm}[ht!]
\caption{Meta-Classification and Fusion Strategy}
\label{alg:meta_fusion}
\begin{algorithmic}[1]
\Require Trained Agent-1 model $f_{\theta_1}$, trained Agent-2 model $f_{\theta_2}$, set of videos $\mathcal{V} = \{V_i\}_{i=1}^N$, ground-truth labels $\{y_i\}_{i=1}^N$
\Ensure Cross-validated $F_1$ score of meta-classifier

\State Initialize empty lists: $\mathcal{P}_1 \gets []$, $\mathcal{P}_2 \gets []$
\For{each video $V_i$ in $\mathcal{V}$}
    \Comment{Agent-1 visual prediction}
    \State Extract up to $m$ frames $\{F_{i,j}\}$ evenly from $V_i$
    \State Normalize and resize each frame
    \State $p^{(1)}_i \gets \frac{1}{|\{F_{i,j}\}|} \sum_{j} f_{\theta_1}(F_{i,j})$
    \State Append $p^{(1)}_i$ to $\mathcal{P}_1$
    \Comment{Agent-2 multimodal prediction}
    \State Extract audio track $A_i$ and transcribe to text $T_a$
    \State Extract visual text $T_v$ from facial regions
    \State Compute MFCC mean vector $\mathbf{a}_i \in \mathbb{R}^{13}$
    \State Compute lexical similarity $s_i = \frac{|\mathcal{W}_a \cap \mathcal{W}_v|}{\max(|\mathcal{W}_a|, 1)}$
    \State Form feature vector $\mathbf{x}_i = [\mathbf{a}_i^\top, s_i]^\top \in \mathbb{R}^{14}$
    \State $p^{(2)}_i \gets f_{\theta_2}(\mathbf{x}_i)$
    \State Append $p^{(2)}_i$ to $\mathcal{P}_2$
\EndFor

\State Form meta-feature matrix $\mathbf{Z} \gets [\mathcal{P}_1, \mathcal{P}_2] \in \mathbb{R}^{N \times 2}$
\State Standardize $\mathbf{Z}$ to zero mean and unit variance per column

\State Initialize meta-classifier $g_{\phi}$ as Random Forest with $T = 100$ trees
\State Apply stratified $K$-fold cross-validation with $K = 5$
\For{each fold}
    \State Split $\mathbf{Z}$ and $y$ into train and validation sets
    \State Train $g_{\phi}$ on training meta-features
    \State Evaluate on validation set and compute $F_1$ score
\EndFor

\State \Return Mean cross-validated $F_1$ score
\end{algorithmic}
\end{algorithm}

\subsubsection{Meta-Feature Fusion and Decision Ensemble}

The set of meta-feature vectors $\{\mathbf{z}_i\}_{i=1}^N$ and their corresponding labels $\{y_i\}_{i=1}^N$ are used to train a second-level classifier. Before training, the way each feature dimension is standardized is provided in Equation \ref{eq:24}:
\begin{equation}
z'_{i,k} = \frac{z_{i,k} - \mu_k}{\sigma_k}, \quad k \in \{1,2\}
\label{eq:24}
\end{equation}
where $\mu_k$ and $\sigma_k$ are the empirical mean and standard deviation of the $k$-th meta-feature across the training set.

The meta-classifier is an ensemble of $T=100$ decision trees, each trained on a bootstrap sample of the training set. For a given input $\mathbf{z}'_i$, each decision tree $h_t$ produces a binary prediction $c_{i,t} \in \{0,1\}$. \(\sum_{t=1}^T\) denotes summation over all \(T\) trees (where \(t\) enumerates individual trees), and the final probability score $\hat{y}$ is the average of all tree outputs, as defined in Equation \ref{eq:25}. Equation \ref{eq:26} represents the predicted class label formulation:
\begin{equation}
\hat{y}^{(\text{meta})}_i = \frac{1}{T} \sum_{t=1}^T h_t(\mathbf{z}'_i)
\label{eq:25}
\end{equation}

\begin{equation}
\tilde{y}_i = \mathbb{I}\left( \hat{y}^{(\text{meta})}_i \geq 0.5 \right)
\label{eq:26}
\end{equation}

Here, \(\mathbb{I}(\cdot)\) is the indicator function that returns \(1\) if its argument is true and \(0\) otherwise. To assess the model's reliability, a stratified $K$-fold cross-validation method is applied, employing $K=5$, while maintaining the balance between actual and forged samples for each fold. The macro-averaged $F_1$ score is computed as defined in Equation \ref{eq:27}:
\begin{equation}
F_1 = \frac{1}{2} \left( \frac{2 \cdot \text{TP}_0}{2 \cdot \text{TP}_0 + \text{FP}_0 + \text{FN}_0} + \frac{2 \cdot \text{TP}_1}{2 \cdot \text{TP}_1 + \text{FP}_1 + \text{FN}_1} \right)
\label{eq:27}
\end{equation}
where $\text{TP}$, $\text{FP}$, and $\text{FN}$ denote the true positives, false positives, and false negatives for class $c \in \{0,1\}$.

This multi-layered decision fusion system ensures that the final classification has access to the visual expertise of Agent-1 and the cross-modal audio-based reasoning ability of Agent-2, and consequently offers improved detection performance for those cases for which one modality may not be sufficient. For clarity and reproducibility, Table \ref{table:x} summarizes the main modules, inputs, and outputs of Agent-1, Agent-2, and the Random Forest meta-classifier in the proposed DeepAgent framework.

\begin{table*}[ht!]
\centering
\scriptsize
\caption{Overview of the DeepAgent architecture, summarizing the main modules of Agent-1 (visual detector), Agent-2 (audio–visual semantic consistency detector), and the Random Forest meta-classifier.}
\renewcommand{\arraystretch}{1.35}
\begin{tabularx}{\textwidth}{
p{2.8cm}
p{2cm}
p{6.5cm}
p{2.3cm}
p{3.5cm}
}
\hline
\textbf{Component} & \textbf{Input} & \textbf{Main operations (key hyperparameters)} & \textbf{Output} & \textbf{Role} \\
\hline

\textbf{Agent-1: Input preprocessing} 
& Video file 
& Resize to 224$\times$224; rescale (1/255); data augmentation (rotation 10°, shifts 0.1, zoom 0.1, brightness [0.9–1.1], horizontal flip); normalization 
& 224$\times$224$\times$3 tensor 
& Standardize resolution and increase robustness via augmentation \\

\textbf{Agent-1: Convolutional feature extractor} 
& 224$\times$224$\times$3 tensor 
& 

\textbf{Block 1}: Conv2D(64, 11$\times$11, stride=4) + ReLU + BatchNorm + MaxPool(3$\times$3, stride=2)

\textbf{Block 2}: Conv2D(128, 5$\times$5, same padding) + ReLU + BatchNorm + MaxPool

\textbf{Block 3}: Conv2D(256, 3$\times$3) + ReLU + BatchNorm

\textbf{Block 4}: Conv2D(256, 3$\times$3) + ReLU + BatchNorm

\textbf{Block 5}: Conv2D(128, 3$\times$3) + ReLU + BatchNorm + MaxPool

GlobalAveragePooling2D
& 
Flattened visual embedding 
& Learn hierarchical spatial features from faces/frames \\

\textbf{Agent-1: Classifier head} 
& Visual embedding 
& Dense(1024) + ReLU + Dropout(0.5) + BatchNorm; Dense(512) + ReLU + Dropout(0.5); Dense(2) + Softmax 
& [$p(\text{real}), p(\text{fake})$] 
& Predict deepfake probability for each frame \\

\textbf{Agent-2: Audio feature extraction} 
& Audio waveform 
& Librosa MFCC extraction (n\_mfcc=13, sr=16000); temporal mean aggregation across frames 
& MFCC feature vector (13 dims) 
& Compact representation of spectral envelope and prosody \\ 

\textbf{Agent-2: Speech transcription} 
& Audio waveform 
& Pretrained Whisper ASR model $\rightarrow$ text transcription 
& ASR transcript (string) 
& Lexical content of spoken audio \\ 

\textbf{Agent-2: Visual text extraction} 
& Video frames 
& Sample every 5th frame; grayscale conversion; EasyOCR text detection (English, GPU-accelerated); text aggregation 
& OCR text embedding 
& Representation of on-screen/subtitle text \\ 

\textbf{Agent-2: Audio-visual synchrony} 
& ASR transcript + OCR text 
& Tokenization $\rightarrow$ set intersection; Jaccard-style similarity 
& Text similarity score
& Measure semantic consistency between audio and visual modalities \\ 

\textbf{Agent-2: Feature fusion} 
& MFCC vector + text similarity 
& Concatenation: [MFCC(13), similarity(1)] $\rightarrow$ combined feature vector (14 dims) 
& Multimodal feature vector (14 dims) 
& Unified representation for classification \\ 

\textbf{Agent-2: Classifier} 
& Multimodal feature vector (14 dims) 
& Dense(128) + ReLU + Dropout(0.2); Dense(64) + ReLU + Dropout(0.2); Dense(32) + ReLU; Dense(1) + Sigmoid; Adam optimizer; binary cross-entropy loss 
& \parbox{2.3cm}{%
[$p(\text{AV consistent}),\\ p(\text{AV inconsistent})$]%
}
& Predict deepfake probability based on audio-visual inconsistency \\ 

\textbf{Meta-classifier (fusion)} 
& Agent-1 \& Agent-2 scores 
& \parbox[t]{6.5cm}{%
Feature vector:
$[p_1(\text{real}), p_1(\text{fake}),$
$p_2(\text{consistent}), p_2(\text{inconsistent})]$
$\rightarrow$ Random Forest with $T$ trees%
}
& Final label: real/ fake 
& Decision-level fusion combining both agents' predictions \\
\\
\hline
\end{tabularx}
\label{table:x}
\end{table*}

\section{Experimental Details}\label{sec4}
In this section, we discuss the dataset preparation process, along with the model’s preprocessing and implementation setup.

\subsection{Data Preparation}
Recent advances in multimodal deepfake detection have highlighted the importance of jointly evaluating visual, audio, and audio–visual consistency patterns. However, existing multimodal systems trained on a single dataset often struggle with cross-dataset generalization, largely due to variations in compression, manipulation pipelines, and identity diversity. Motivated by this, our data preparation and training strategy explicitly prioritizes cross-dataset stability.
\subsubsection{Dataset}
Effective evaluation of deepfake detection methods requires datasets that encompass diverse manipulation techniques across both visual and audio domains. In our study, We utilize three popular deepfake datasets: Celeb-DF \cite{li2020celebdf}, FakeAVCeleb \cite{khalid2021fakeavceleb}, and DeepFakeTIMIT \cite{korshunov2018deepfaketimit}. Our evaluation and training strategy is designed to ensure both cross-dataset generalizability and robustness in multitask detection. Specifically, Agent-1 is trained from a combination of FakeAVCeleb and Celeb-DF based on the complementarity of strengths that exist in advanced face interchange methods. In contrast, Agent-2 is trained with FakeAVCeleb due to its unique use of audio-video manipulations. The proposed meta-model has been tested and evaluated on FakeAVCeleb and DeepFakeTIMIT, respectively.

\textbf{Celeb-DF: }Celeb-DF \cite{li2020celebdf} consists of 590 real videos and 5,639 fake videos generated through advanced face swapping techniques applied over the faces of celebrities, and collected from publicly available YouTube clips of 59 celebrities (both male and female) with diverse origins, lighting conditions, and head poses. Each video averages 13 seconds in duration with a 30 frames-per-second (fps) rate and 720p resolution that totals over 2 million frames. This dataset corrects quality gaps found in existing benchmarks by providing more complex and natural samples. 

\textbf{FakeAVCeleb: }FakeAVCeleb \cite{khalid2021fakeavceleb} dataset represents a total of 20,000 videos that consist of 500 real and 19,500 fake video samples, collected from the VoxCeleb2 \cite{chung2018voxceleb2} celebrity dataset. Each real video has an average duration of 7.8 seconds, and the participants cover four major ethnic groups: Caucasian, Black, South Asian, and East Asian, with an equal representation of male and female subjects to reduce bias. 

\textbf{DeepFakeTIMIT: }The DeepFakeTIMIT \cite{korshunov2018deepfaketimit} dataset is one of the first publicly available databases containing deepfake videos generated using a GAN-based face swapping method. It is built from the VidTIMIT dataset, which includes 43 subjects recorded under controlled conditions, and from these, 16 visually similar pairs were selected to perform bidirectional face swaps. Finally, for each pair, two quality versions were produced, named Low Quality (LQ) and High Quality (HQ), and a total of 640 deepfake videos (320 per version).

The summary of datasets used in this study is presented in Table \ref{tab:datasets}.

\begin{table}[ht!]
\centering
\caption{Datasets employed in this study, with details on the number of real and fake samples }
\scriptsize
\begin{tabularx}{\columnwidth}{p{2.2cm} p{1.7cm} p{1.7cm} p{1.9cm}}
\hline
\textbf{Dataset Name} & \textbf{Real Videos} & \textbf{Fake Videos} & \textbf{Fake Audio} \\
\hline
Celeb-DF \cite{li2020celebdf} & 590 & 5639 & No \\
FakeAVCeleb \cite{khalid2021fakeavceleb} & 500 & 19500 & Yes \\
DeepFakeTIMIT \cite{korshunov2018deepfaketimit} & 0 & 640 & Yes \\
\hline
\end{tabularx}
\label{tab:datasets}
\end{table}

\subsubsection{Data Preprocessing}
\textbf{Visual Preprocessing:} The video data are segmented into individual samples, which serve as the primary input samples for Agent-1. To maintain spatial uniformity across all inputs, each video is resized to a fixed resolution of $224 \times 224$ pixels, which results in a standardized tensor for each video. Furthermore, pixel intensity values are normalized to the continuous range $[0,1]$ by dividing the raw values by the maximum intensity of $255$. This normalization procedure stabilizes the learning process and mitigates scale related biases.

The processed video data are grouped into mini-batches of size 16. To reduce computational cost while keeping the temporal flow of the video, a uniform subsampling method is used that selects frames at a regular interval of five. Each video is then converted to grayscale to minimize data redundancy while retaining key spatial and temporal details important for text detection. Finally, an OCR module analyzes these grayscale video sequences to extract the visual text information across time. We convert videos to grayscale, so color information is removed. In our datasets, most deepfake artifacts are caused by shape and texture problems, such as misaligned facial features, overly smooth skin, or jerky and inconsistent motion between frames. These issues do not depend strongly on colour. Using only a grayscale (luminance) channel makes the input smaller, training more stable, and memory use lower, which is important for Agent-1’s lightweight design. In our initial tests with RGB frames, accuracy did not consistently improve. For this reason, we use grayscale input as a more efficient choice that still preserves the key visual artifacts needed for detection.

\textbf{Audio-Text Preprocessing:} The first step is the extraction of the audio stream from the video file. Then the waveform goes through processing and resampling, where MFCCs are computed. To reduce temporal variability, the temporal mean of the MFCC features is calculated, resulting in compact audio embeddings. At the same time, the audio is also transcribed using the Whisper model to obtain textual versions of the spoken content. In order to attain modality alignment, the transcribed audio text is measured against the text from the video by lexical overlap, thereby providing a similarity score that reflects semantic similarity across the two visual and auditory modalities. Collectively, these operations ensure spatial uniformity, temporal efficiency, and informative feature representations, which are subsequently employed for multimodal feature extraction and analysis. 

Agent-1 is designed as a purely visual detector, so it is trained on Celeb-DF and the visual stream of FakeAVCeleb, which together provide high-quality, high-variability face videos under diverse compression levels and manipulation pipelines. Celeb-DF in particular is known for its visually challenging deepfakes with fewer obvious artifacts, while FakeAVCeleb offers additional variety in terms of identity and pose. Agent-2, in contrast, requires tightly synchronized audio–video pairs and reliable speech content to compute audio–visual semantic consistency. For this reason, it is trained and evaluated primarily on FakeAVCeleb and DeepFakeTIMIT, where every clip includes aligned speech and footage, enabling robust extraction of MFCCs, Whisper transcripts, and OCR text. This division ensures that each agent is exposed to the type of data most relevant to its specialization, while the meta-classifier later combines their complementary strengths.

\subsection{Implementation Setup}
This research is conducted using an AMD Ryzen 7700 8-core Central Processing Unit (CPU) operating at 5.3 GHz and 16 GB of RAM for all experiments. For graphical processing, an NVIDIA GeForce RTX 4060 with 8 GB of VRAM is utilized. All computations are executed on a Windows 11 platform, while Jupyter Notebook version 7.0.8 serves as the integrated development environment (IDE). All models are implemented in Python 3.10 using TensorFlow 2.x and Keras for deep learning components.

For Agent-1, we have used the Adam optimizer with initial learning rate $\alpha=0.0001$, exponential decay rates $\beta_1=0.9$ (first moment) and $\beta_2=0.999$ (second moment), and numerical stability constant $\epsilon=10^{-7}$, training for up to 50 epochs with batch size 16. Agent-2 is trained using Adam optimizer (default TensorFlow parameters) for 100 epochs with a batch size of 16. Early stopping with patience of 10 epochs and learning rate reduction (factor=0.5, patience=5) is applied based on validation accuracy. The dataset is split into training (70\%), validation (20\%), and test (10\%) subsets with a fixed random seed of 42 for reproducibility across all splits. The meta-classifier is employed 5-fold stratified cross-validation to ensure robust performance estimation across different data partitions.

\section{Result Analysis}\label{sec5}
This section presents the experimental outcomes of the DeepAgent framework. The performance of Agent-1 and Agent-2 is evaluated individually, and the fused predictions derived from the meta-classifier are also shown. Classification accuracy, precision, recall, F1-score, and overall robustness are analyzed on benchmark datasets.

\subsection{Performance of the Proposed Framework}
To understand the effectiveness of DeepAgent, we have analysed the performance of each agent on its respective modality of specialisation and the improvement of the fusion of its results on the dataset stability of performance. The findings indicate that visual features and audio-text-visual consistency give complementary information, which is more effective to use to make the model more reliable in distinguishing real and manipulated videos.

\subsubsection{Performance of Individual Agent}
The performance evaluation of Agent-1 demonstrates its effectiveness in detecting deepfakes using only visual features extracted from videos through a lightweight CNN-based architecture. The detailed performance metrics are summarized in Table~\ref{tab:individual_agent}. Agent-1 obtains an overall accuracy of 90\% and a test accuracy of 94.35\% on the combined Celeb-DF and FakeAVCeleb dataset. The model performed exceptionally well on the Real class, with a precision, recall, and F1-score of 91\%, 96\%, and 93\%, respectively. The Fake class performed worse, with a recall of 74\% and an F1-score of 80\%. These metrics suggest that whereas Agent-1 is very consistent in detecting real videos, it sometimes fails to catch fine manipulations in highly compressed or low resolution deepfakes, where inconsistencies at the pixel level are harder to identify. 

Agent-2 combines MFCC-based audio embeddings, Whisper generated \cite{radford2023robust} transcripts, and OCR extracted text features to assess the semantic consistency between audio and visual modalities, with its performance summarized in Table \ref{tab:individual_agent}. On the FakeAVCeleb dataset, Agent-2 obtains an overall accuracy of 86\% and a test accuracy of 93.69\%. The model demonstrates balanced performance across classes, with the Fake class achieving a precision of 90\% and a recall of 83\% (F1-score 86\%), and the Real class obtains a precision of 82\% and a recall of 89\% (F1-score 86\%). These findings indicate that Agent-2 excels at detecting cross-modal inconsistencies, including misalignment between spoken words and lip movement. 


\begin{table}[ht!]
\centering
\caption{Classification metrics for Agent-1 and Agent-2}
\scriptsize
\begin{tabular}{lcc|cc}
\toprule
 & \multicolumn{2}{c}{\textbf{Agent-1 (\%)}} & \multicolumn{2}{c}{\textbf{Agent-2 (\%)}} \\ 
\bottomrule
\textbf{Metric}  & \textbf{Fake} & \textbf{Real} & \textbf{Fake} & \textbf{Real} \\ 
Precision     & 86 & 91 & 90 & 82 \\ 
Recall        & 74 & 96 & 83 & 89 \\ 
F1-score      & 80 & 93 & 86 & 86 \\ 
Macro Avg     & 85 & 89 & 86 & 86 \\ 
Weighted Avg  & 90 & 90 & 86 & 86 \\ 
Train Accuracy      & \multicolumn{2}{c}{90} & \multicolumn{2}{c}{86} \\ 
Test Accuracy & \multicolumn{2}{c}{94.35} & \multicolumn{2}{c}{93.69} \\ 
\bottomrule
\end{tabular}
\label{tab:individual_agent}
\end{table}

The ablation results in Table \ref{tab:modality_comparison_agent2} clearly show how each component of Agent-2 contributes to its final performance. With the help of MFCC-based audio features alone, the accuracy is 87.94\%, which indicates that these 13-dimensional audio indicators retain valuable hints introduced by the synthetic speech generation. When the Whisper generated transcripts are incorporated with the MFCC features, the accuracy is further enhanced to 89.45\%, indicating that the spoken words can also be used to aid the model in identifying semantic irregularities, particularly when the audio features cannot be used solely to identify them. When the three modalities, namely, MFCCs, transcripts, and OCR extracted frame text, are combined as the final 14-dimensional representation, the highest accuracy is reached, 93.69\%. In general, the increasing trend of accuracy indicates that all modalities contribute to the useful, non-overlapping information and prove the effectiveness of the multimodal fusion strategy of Agent-2.

\begin{table}[ht!]
\centering
\caption{Accuracy comparison using different modality combinations on Agent-2}
\scriptsize
\begin{tabular}{cccc}
\toprule
\textbf{Audio (MFCC)} & \textbf{Transcript} & \textbf{OCR} & \textbf{Accuracy (\%)} \\
\midrule
\cmark & \xmark & \xmark & 87.94 \\
\cmark & \cmark & \xmark & 89.45 \\
\cmark & \cmark & \cmark & 93.69 \\
\bottomrule
\end{tabular}
\label{tab:modality_comparison_agent2}
\end{table}
\begin{table*}[ht!]
\centering
\caption{Performance of the meta-classifier, evaluated on the \textbf{FakeAVCeleb} dataset using 5-fold cross-validation.}
\scriptsize
\begin{tabular}{c c c c c c}
\toprule
\textbf{Fold} & \textbf{Accuracy (\%)} & \textbf{Precision (\%)} & \textbf{Recall (\%)} & \textbf{F1 Score (\%)} & \textbf{AUC (\%)} \\
\bottomrule
1 & 76.67 & 79.27 & 72.22 & 75.58 & 87.39 \\
2 & 82.22 & 79.00 & 87.78 & 83.16 & 91.51 \\
3 & 87.78 & 92.50 & 82.22 & 87.06 & 93.64 \\
4 & 78.89 & 77.08 & 82.22 & 79.57 & 86.15 \\
5 & 82.22 & 81.52 & 83.33 & 82.42 & 89.59 \\
\textbf{Mean} & 81.56 & 81.87 & 81.56 & 81.56 & 89.66 \\
\bottomrule
\end{tabular}
\label{tab:meta_classifier_results}
\end{table*}

\subsubsection{Meta-classifier Performance}
We fuse the predictions from both agents using a Random Forest meta-classifier. On FakeAVCeleb, it achieves a mean accuracy of 81.56\%, precision of 81.87\%, recall of 81.56\%, and an F1-score of 81.56\%, demonstrating more stable and generalized performance across folds. The results shown in Table~\ref{tab:meta_classifier_results} indicate improvements across all evaluation metrics.

\textbf{1. Confusion Matrix: }
Figure \ref{fig:confusion} presents the confusion matrices of DeepAgent, obtained from a 5-fold stratified cross-validation. Figure \ref{fig:confusion} (A) corresponds to results on the FakeAVCeleb dataset, while Figure \ref{fig:confusion} (B) illustrates performance on the DeepFakeTIMIT dataset. The diagonal elements in each matrix represent the correctly classified samples and demonstrate the ability of the proposed model to consistently distinguish between real and fake instances. For FakeAVCeleb, a few misclassifications are visible across folds, yet the overall pattern indicates strong predictive consistency, with most samples correctly labeled. Despite these few misclassifications, the majority of samples are correctly identified, and the model has achieved an overall accuracy of 81.56\% on the FakeAVCeleb dataset and 97.49\% on the DeepFakeTIMIT dataset.

\begin{figure*}[ht!]
\centering
\includegraphics[scale=0.3]{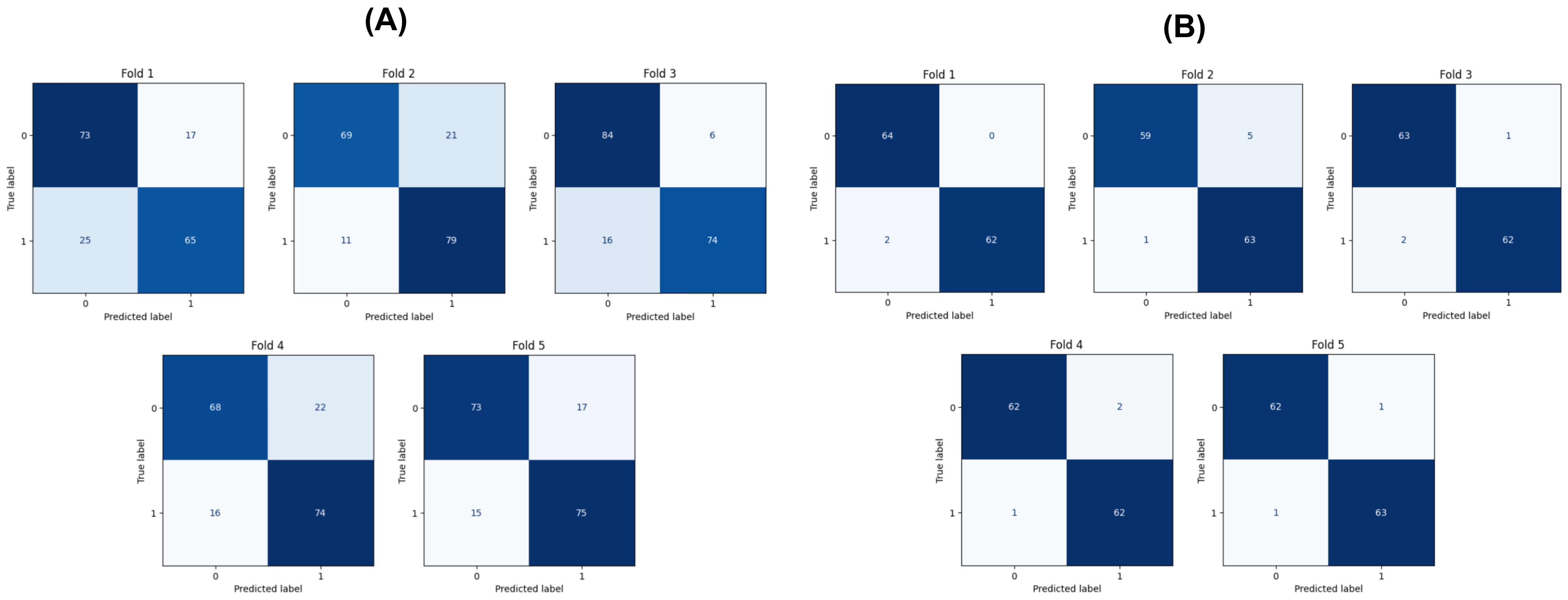}
\caption{Confusion matrices of DeepAgent obtained from 5-fold stratified cross-validation on (A) FakeAVCeleb and (B) DeepFakeTIMIT datasets.}
\label{fig:confusion}
\end{figure*}

\textbf{2. Receiver Operating Characteristic Curve Analysis: }
The Receiver Operating Characteristic (ROC) curves generated from our proposed DeepAgent model’s predictions on both datasets are presented in Figure \ref{fig:roc}. The curve shown in Figure \ref{fig:roc} (A) illustrates the model’s performance on the FakeAVCeleb dataset, whereas Figure \ref{fig:roc} (B) denotes the corresponding results on the DeepFakeTIMIT dataset. The ROC curve explains the relationship between the true positive rate and the false positive rate, and offers a reliable measure of the model’s ability to differentiate between real and fake instances. The model achieves a mean AUC of 0.897 on the FakeAVCeleb dataset, that indicates its reliable classification ability. Obtaining a nearly perfect mean AUC of 0.998 on the DeepFakeTIMIT dataset highlights the stable detection performance of DeepAgent, with very limited overlap between real and fake classes, and confirms the model’s robustness and high generalization capability across datasets.

\begin{figure*}[ht!]
\centering
\includegraphics[scale=0.45]{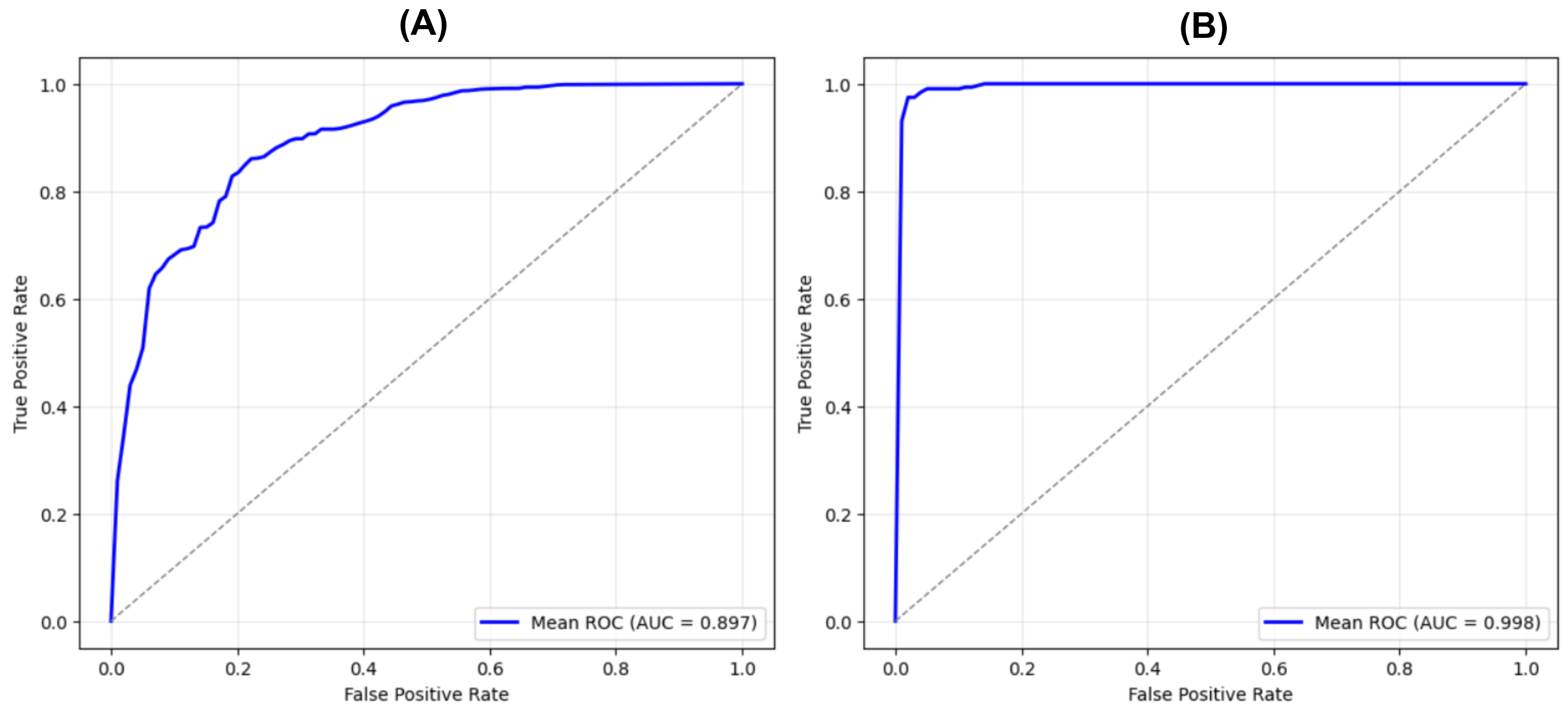}
\caption{ROC curves illustrating the performance of the proposed DeepAgent model on (A) FakeAVCeleb dataset and (B) DeepFakeTIMIT dataset}
\label{fig:roc}
\end{figure*}

\subsection{Cross Dataset Validation}
To comprehensively assess the robustness and consistency of the proposed DeepAgent framework, we performed cross-dataset evaluation on the DeepFakeTIMIT dataset. The model demonstrates consistently high and stable performance across all folds, achieving an average accuracy of 97.49\%, precision of 97.27\%, recall of 97.81\%, F1-score of 97.52\%, and AUC of 99.75\%. These results indicate that DeepAgent effectively balances sensitivity and precision and maintains excellent discriminative capability across diverse video samples. 

Across the five folds of cross-validation on the DeepFakeTIMIT dataset, the performance of DeepAgent has remained consistently high with only minor variations. In the first fold, DeepAgent achieves the highest accuracy of 98.44\% and a perfect precision that demonstrates its strong ability to identify fake instances correctly. The second fold yields a slightly lower accuracy of 95.31\%; however, it maintains a high recall of 98.44\%, indicating reliable detection sensitivity. Similarly, the third and fourth folds produce balanced outcomes with accuracies of 97.66\% and 97.64\%, respectively. The fifth fold achieves 98.43\% accuracy, with both precision and recall at 98.44\%, confirming excellent classification consistency, as summarized in Table \ref{tab:deepfake_timit_results}.

\begin{table*}[ht!]
\centering
\caption{Performance of the meta-classifier, evaluated on the \textbf{DeepFakeTIMIT} dataset using 5-fold cross-validation.}
\scriptsize
\begin{tabular}{c c c c c c}
\toprule
\textbf{Fold} & \textbf{Accuracy (\%)} & \textbf{Precision (\%)} & \textbf{Recall (\%)} & \textbf{F1 Score (\%)} & \textbf{AUC (\%)} \\
\bottomrule
1 & 98.44 & 100.00 & 96.88 & 98.41 & 99.95 \\
2 & 95.31 & 92.65 & 98.44 & 95.45 & 99.56 \\
3 & 97.66 & 98.41 & 96.88 & 97.64 & 99.73 \\
4 & 97.64 & 96.88 & 98.41 & 97.64 & 99.75 \\
5 & 98.43 & 98.44 & 98.44 & 98.44 & 99.78 \\
\textbf{Mean} & 97.49 & 97.27 & 97.81 & 97.52 & 99.75 \\
\bottomrule
\end{tabular}
\label{tab:deepfake_timit_results}
\end{table*}

\section{Discussion}\label{sec6}
Our proposed DeepAgent framework introduces a multi-agent deepfake detection system that jointly analyzes audio-visual modalities to identify deepfake video content. The architecture is organized into two complementary detection agents and a meta-fusion layer that collaboratively enhance robustness and generalization. Agent-1 specializes in visual feature learning through a lightweight CNN, optimized to detect spatial and temporal artifacts in videos. It extracts local and global facial features through hierarchical convolution and pooling, and forms a low dimensional discriminative embedding using global average pooling followed by dropout regularized dense layers. This modular design allows the model to encode visual distortions and compression artifacts commonly present in synthetic content. Additionally, Agent-2 serves as an audio-visual semantic consistency detector, integrating both speech and visual cues to identify cross-modal inconsistency. It extracts MFCCs from the audio signal, textual information from video frames using OCR, and transcribes speech content. A cross-modal similarity score quantifies the semantic alignment between the visual and spoken text components, which is then concatenated with the audio embeddings to construct a unified 14-dimensional multimodal representation. The resulting representation allows Agent-2 to identify semantic desynchronization.

To combine the outputs from both agents, DeepAgent uses a meta classification and fusion strategy built on a Random Forest ensemble. The prediction scores from Agent-1 and Agent-2 are joined together to form a single meta-feature vector, which is then given as input to the ensemble classifier. Each tree in the Random Forest makes its own decision, and the average of all these decisions gives the final probability that a video is a deepfake. This layered fusion approach allows the model to make use of information from both the audio-visual domains while reducing the weaknesses that may exist in each agent. This combination of dual specialized agents: one capturing fine grained visual artifacts and the other enforcing audio-visual semantic consistency with MFCC-based audio embeddings, together with a meta-fusion layer, represents a distinctive approach in deepfake detection. To ensure a fair and reliable evaluation, the system also uses stratified cross-validation, which keeps a balance between real and fake samples.

The proposed DeepAgent framework demonstrates strong and consistent performance across multiple benchmark datasets and confirms its reliability and robustness in deepfake detection. In the individual evaluation, Agent-1 achieved a test accuracy of 94.35\%, effectively detecting visual inconsistencies within manipulated videos. Its performance on the Real class has been particularly strong, with a precision of 91\%, recall of 96\%, and F1-score of 93\%, reflecting its high capability in recognizing authentic video samples. However, the Fake class achieved a slightly lower recall of 74\% and F1-score of 80\%. The training and validation curves indicate consistent improvement across epochs without signs of overfitting. Agent-2 also shows promising performance with a test accuracy of 93.69\%. By combining MFCC-based acoustic embeddings, Whisper transcribed speech, and OCR derived text, it achieved balanced class detection. The Fake class obtained precision and recall values of 90\% and 83\% (F1-score 86\%), while the Real class reached a precision of 82\%, recall 89\%, and F1-score 86\%. These results highlight Agent-2’s ability to detect semantic inconsistencies between audio and visual cues. Although its precision was slightly lower than Agent-1 due to noisy audio samples, the model has demonstrated steady learning behavior and avoided overfitting throughout training.

Compared to recent existing models, our proposed DeepAgent framework demonstrates superior performance and consistency across benchmark datasets. It achieves a mean accuracy of 97.49\% on the DeepFakeTIMIT dataset and 81.56\% on FakeAVCeleb, outperforming several established deepfake detection approaches. The difference in performance between FakeAVCeleb and DeepFakeTIMIT reflects the natural variation in dataset characteristics. The FakeAVCeleb dataset contains more diverse identities, environments, and manipulation styles, whereas DeepFakeTIMIT is more uniform and controlled, allowing DeepAgent to operate closer to optimal performance. Traditional hybrid architectures and representation learning–based approach, such as OF+CNN+LSTM \cite{saikia2022hybrid}, RFFR \cite{shi2025real}, and YOLO-CNN-XGBoost \cite{ismail2021new} reported slightly lower accuracies of 79.49\%, 88.98\%, and 90.73\%, respectively. In contrast, transformer-based models, such as DeepFake-Adapter \cite{shao2025deepfake} and STKD-VViT \cite{usmani2025spatio} achieved up to 96.52\% and 96.00\%. Similarly, C2P-CLIP \cite{tan2025c2p} and Hybrid CNN+CViT \cite{soudy2024deepfake} reached accuracies of 95.80\% and 97.00\%, yet fell slightly short of DeepAgent’s performance, highlighting this framework's advantage in integrating multimodal reasoning and meta-fusion strategies, which enhance both precision and robustness.

While our proposed DeepAgent demonstrates promising results, it has a few limitations. Firstly, although late fusion stabilizes outputs, it fails to dynamically respond to those situations where a modality is more trustworthy than the other. An adaptive fusion scheme, possibly guided by uncertainty estimation, may further enhance decision quality. Secondly, inference efficiency is limited by the requirement to process both modalities and may inhibit scalability in situations where real time or compute limited conditions apply. Thirdly, our evaluation covered multiple datasets; however, future work should focus on using larger and more diverse multimodal datasets to ensure fair performance across different demographic groups and to improve robustness against adversarial manipulations. 

\section{Conclusion}\label{sec7}
Our work introduces and evaluates DeepAgent, an innovative multi-agent cooperative framework that integrates visual and audio intelligence to enhance the robustness and reliability of deepfake detection. Agent-1, based on a lightweight CNN, achieved a test accuracy of 94.35\% on the combined Celeb-DF and FakeAVCeleb datasets, while Agent-2 captures audio-visual inconsistencies with 93.69\% accuracy on the FakeAVCeleb dataset. The framework of combining the agents achieved an accuracy of 81.56\% on the FakeAVCeleb dataset and 97.49\% accuracy on the DeepFakeTIMIT dataset.

Our work presents an advanced architecture composed of two collaborative agents to enhance the accuracy and reliability of deepfake detection. Agent-1 focuses on visual analysis using a lightweight CNN to identify pixel-level inconsistencies, while Agent-2 detects audio-visual mismatches by combining acoustic features, Whisper-based transcriptions, and OCR extracted frame text. The agents are integrated through a meta-classifier, which effectively improves overall performance. The model is evaluated on the Celeb-DF, FakeAVCeleb, and DeepFakeTIMIT datasets to assess its generalization ability across diverse manipulations. All of these, together, contribute to the model's strong detection capability. 

The datasets used in this work are largely celebrity focused and may not reflect the full range of age, gender, ethnicity, and speaking styles encountered in practice. We do not currently report subgroup performance or analyze potential biases in false positive and false negative rates across demographic groups. Furthermore, the deployment context and threat model are only partially explored. Future work should incorporate fairness aware evaluation, stratified analysis across demographic attributes where available, and more explicit modeling of the intended application scenarios and attacker capabilities. With its ability to capture both audio-visual inconsistencies, DeepAgent has strong potential to strengthen digital trust and improve the reliability of automated verification systems.

\section*{Statements and Declarations}

\textbf{Funding:} This study does not include any external funding.\\
\textbf{Ethical Approval:} Not applicable\\
\textbf{Competing interests:} The authors state that they have no known financial conflicts of interest or personal relationships that could have influenced the work presented in this paper.\\ 
\textbf{Data Availability:} In this study we have used three publicly available datasets, including \href{https://ieeexplore.ieee.org/document/9156368}{Celeb-DF} \cite{li2020celebdf}, \href{https://arxiv.org/abs/2108.05080}{FakeAVCeleb} \cite{khalid2021fakeavceleb}, and \href{https://www.idiap.ch/en/scientific-research/data/deepfaketimit}{DeepfakeTIMIT} \cite{korshunov2018deepfaketimit}.
\textbf{CRediT authorship contribution statement:}

\textbf{Sayeem Been Zaman}: Conceptualization, Literature review, Formal analysis Methodology, Writing: original draft, Validation, Writing – review \& editing;
\textbf{Wasimul Karim}: Conceptualization, Literature review, Formal analysis Methodology, Writing: original draft, Validation, Writing – review \& editing; 
\textbf{Arefin Ittesafun Abian}: Conceptualization, Literature review, Formal analysisSupervision, Project administration, Writing – original draft, Writing – review \& editing; 
\textbf{Reem E. Mohamed}: Validation, Formal analysis, Writing – review \& editing;
\textbf{Md Rafiqul Islam}: Validation, Formal analysis, Writing – review \& editing;
\textbf{Asif Karim}: Validation, Formal analysis, Writing – review \& editing;
\textbf{Sami Azam}: Conceptualization, Supervision, Project administration, Validation, Writing: review.

\end{document}